\documentclass[conference]{IEEEtran}
\IEEEoverridecommandlockouts
\usepackage{cite}
\usepackage{amsmath,amssymb,amsfonts}
\usepackage{algorithmic}
\usepackage{graphicx}
\usepackage{textcomp}
\usepackage{xcolor}
\usepackage{booktabs}
\usepackage{threeparttable}
\usepackage{orcidlink}
\usepackage{booktabs} 
\usepackage{multirow} 
\usepackage{threeparttable} 
\usepackage{pifont} 
\usepackage{hyperref}
\providecommand{\cmark}{\ding{51}} 
\providecommand{\xmark}{\ding{55}}

\def\BibTeX{{\rm B\kern-.05em{\sc i\kern-.025em b}\kern-.08em
    T\kern-.1667em\lower.7ex\hbox{E}\kern-.125emX}}
\begin{document}

\title{LEViL: Label-Efficient Video Learning via Zero-Shot Distillation over VLM-Generated Pseudo-Label Spaces\\

\thanks{This work was supported by the Scientific and Technological Research Council of Türkiye (TÜBİTAK) under Grant No.~125E351 and by the Kocaeli University Scientific Research Projects Coordination Unit under Grant No.~4531.}
}

\author{
\IEEEauthorblockN{
Aslı Çelik\,\orcidlink{0000-0003-1874-6743}
}
\IEEEauthorblockA{
\textit{Department of Electronics and Communications Engineering}\\
\textit{Kocaeli University}\\
Kocaeli, Türkiye\\
asli.celik@kocaeli.edu.tr
}
}

\maketitle

\begin{abstract}
Supervised video pretraining is a common transfer learning practice for improving downstream action recognition performance. However, it requires large-scale labeled source datasets, and the effectiveness of the learned initialization is influenced by the similarity between the source and target domains. Constructing such labeled pretraining datasets for different target domains is costly and difficult to scale. To address these limitations, this study proposes a label-efficient video learning framework that combines annotation-free video pretraining with target-label-set-aware fine-tuning. During pretraining, a vision-language model (VLM) generates textual descriptions of unlabeled videos, which are processed to construct an interpretable semantic pseudo-label space. A frozen video-language model then produces zero-shot soft target distributions over this space, allowing a student video encoder to learn semantically rich representations without manual source annotations. During downstream adaptation, target-label-set-aware fine-tuning combines supervised learning from labeled target videos with zero-shot distillation over the actual target label set, helping preserve VLM-derived semantic guidance while adapting the pretrained encoder to the target task. Experiments on UCF101 and HMDB51 show that the proposed framework outperforms the compared semi-supervised video action recognition methods across all evaluated limited-label regimes. Moreover, the annotation-free pretraining stage learns transferable representations that provide an effective initialization for full-data fine-tuning, despite relying on a comparatively modest unlabeled pretraining pool.
\end{abstract}

\section{Introduction}
\label{sec:introduction}
Training deep learning models on video data remains challenging because manual video annotation is costly, and supervised learning with limited labeled data can easily lead to overfitting and poor generalization. Compared with image-based models, video models are often more computationally demanding and may contain more parameters due to the additional temporal dimension, increasing their need for training data. Transfer learning is widely used to address these challenges by leveraging representations learned through pretraining on large-scale source datasets. In video learning, this is commonly achieved through supervised pretraining on large-scale video datasets, such as Kinetics~\cite{kinetics} or Sports-1M~\cite{sports1m}. Although supervised video pretraining is a well-established strategy for improving downstream performance, it still depends on manually annotated source videos, and the effectiveness of the learned representations may vary according to the alignment between the source and target domains. Moreover, constructing large-scale labeled video datasets is neither scalable nor sustainable.

Beyond the need for labeled source datasets, supervised video pretraining is also constrained by its reliance on predefined action categories. Models trained with cross-entropy receive supervision through one-hot labels, which provide only a coarse description of the video content. A single action label does not explicitly encode information about the objects, scenes, interactions, and temporal context present in the video. When labeled data are limited, such coarse supervision may be insufficient for learning the broader semantic structure of video content, motivating the use of more informative supervision signals.

Several alternatives to fully supervised video pretraining have been explored to reduce reliance on manually labeled source datasets. Existing transfer learning strategies for video models include 2D-to-3D weight inflation, supervised video pretraining, knowledge distillation from teacher models, and self-supervised pretraining~\cite{aslitransfer}. Among these, knowledge distillation is particularly relevant to the present work, as it enables a student video model to learn from soft supervision produced by a stronger teacher~\cite{hinton}. However, conventional distillation-based initialization methods for video often rely on image-level teachers and transfer knowledge through fixed outputs or predefined label spaces, limiting their ability to exploit the richer semantic information available in unlabeled videos\cite{danet,distinit}.

Recent progress in visual representation learning offers a promising way to address these limitations. The field has increasingly shifted from task-specific models toward foundation models that can be adapted to downstream tasks with limited supervision. Within this broader paradigm, vision-language models (VLMs) are particularly relevant because they align visual and textual representations within a shared semantic embedding space, enabling semantic supervision that extends beyond fixed category labels~\cite{clip,align}. Owing to their large-scale multimodal pretraining, VLMs have demonstrated strong zero-shot and fine-tuning performance across a broad range of visual recognition and retrieval tasks, and this paradigm has subsequently been extended from images to video~\cite{clip,siglip,frozen,cliphiker,vificlip,vljepa}. More recently, the integration of visual encoders with large language models (LLMs) has expanded VLM capabilities beyond discriminative recognition toward open-ended visual-language understanding, including captioning, visual reasoning, question answering, and interactive dialogue~\cite{flamingo,qwenvl,kosmos,instructblip}. In the video domain, such models can generate descriptions that capture not only actions but also objects, scenes, interactions, and temporal context~\cite{internvideo,videochat,videochatgpt,videollama}. These capabilities make them promising sources of rich semantic supervision for video pretraining and representation learning.

Motivated by these observations, this study proposes an annotation-free video pretraining method based on zero-shot distillation over a VLM-generated pseudo-label space, together with a target-label-set-aware fine-tuning strategy for downstream adaptation. A VLM first generates textual descriptions of unlabeled videos, which are then processed using a natural language processing (NLP) pipeline to construct an automatic pseudo-label space from the semantic information discovered in the videos. Unlike conventional label spaces defined by a single action category, the resulting space captures multiple semantic concepts, including actions, objects, and action--object interactions. A frozen video-language model then produces zero-shot soft target distributions over this space based on video--text similarities, and a student video encoder is pretrained by distilling these targets. When labeled target videos are available, the pretrained encoder is further adapted through target-label-set-aware fine-tuning, where supervised learning with ground-truth labels is combined with zero-shot distillation over the actual target label space. The proposed framework is evaluated on downstream video action recognition under both limited-label fine-tuning and full-data fine-tuning settings.

The main contributions of this work are summarized as follows:
\begin{itemize}

\item This work proposes an annotation-free video pretraining framework that learns transferable video representations from unlabeled videos, providing a scalable alternative to supervised pretraining on labeled source datasets and hand-designed self-supervised pretext tasks.

\item A VLM-guided pseudo-label space construction strategy derives an interpretable semantic vocabulary from VLM-generated video descriptions, enabling supervision beyond predefined action categories.

\item A zero-shot soft distillation objective is formulated over the generated pseudo-label space using a frozen video-language model, allowing a student video encoder to learn from soft target distributions without source annotations.

\item A target-label-set-aware fine-tuning strategy combines supervised learning from labeled target videos with zero-shot distillation over the target label space for downstream action recognition.

\end{itemize}

\section{Related Work}
\label{sec:relatedwork}
\subsection{Transfer Learning for Video Models}

Transfer learning is widely used for video models, particularly when labeled target data are limited. Existing approaches obtain model initialization through several mechanisms, including 2D-to-3D weight inflation, supervised video pretraining, knowledge distillation, and self-supervised pretraining~\cite{aslitransfer}. Supervised pretraining on large-scale datasets such as Kinetics remains a strong baseline, whereas self-supervised methods learn from unlabeled videos using pretext-based~\cite{speednet,3drotnet,3dstpuzzle,cliporder}, contrastive~\cite{videomoco,tclr}, or generative~\cite{videomae,motionmae} objectives. Knowledge distillation provides another direction by transferring supervisory information from a pretrained teacher to a student video model~\cite{distinit,danet}. Among these approaches, the proposed method is most closely related to distillation-based transfer learning for video models. DistInit~\cite{distinit} demonstrated that video representations can be learned from image-pretrained teachers without labeled video data, while subsequent methods extended distillation-based initialization to limited-label video learning by using teacher predictions as auxiliary supervision~\cite{videossl,danet,ours}. These approaches typically transfer knowledge through teacher features or predictions defined over fixed label spaces. In contrast, the proposed method constructs an interpretable pseudo-label space directly from VLM-generated descriptions of unlabeled videos and performs zero-shot distillation over this derived space.

\subsection{Vision-Language Models}

In recent years, vision-language models (VLMs) have attracted considerable attention due to their strong performance across a wide range of visual-language tasks. These models commonly learn to align visual and textual modalities within a shared semantic embedding space. Early models such as CLIP~\cite{clip} and ALIGN~\cite{align} demonstrated that large-scale image--text contrastive learning enables strong zero-shot recognition by comparing visual embeddings with natural-language descriptions. This paradigm has also been extended to the video domain~\cite{frozen,cliphiker,vificlip}.

Beyond contrastive image--text and video--text representation learning, the integration of visual encoders with large language models (LLMs) has expanded VLMs toward generative and instruction-following multimodal systems. These models, often referred to as vision-language large models (VLLMs) or multimodal large language models (MLLMs), can generate textual descriptions, answer visual questions, and perform open-ended visual-linguistic reasoning~\cite{flamingo,qwenvl,kosmos,instructblip,internvl}. These capabilities have also been extended to the video domain, where video-centric multimodal models process temporal visual inputs to support open-ended video description, question answering, and dialogue~\cite{internvideo,videochat,videochatgpt,videollama}. Such models therefore provide a promising source of semantic supervision for video representation learning, as they can identify actions, objects, and interactions beyond a predefined set of action categories.

Building on these capabilities, the proposed method performs annotation-free video pretraining by first generating textual descriptions for unlabeled videos using a VLM and then transforming these descriptions into a structured semantic pseudo-label space. A frozen video-language model is subsequently employed to produce zero-shot soft target distributions over the resulting textual vocabulary. Unlike approaches that use VLMs solely as zero-shot classifiers or feature teachers, the proposed framework exploits them at two complementary stages: first, to construct an interpretable supervision space directly from unlabeled video content, and second, to provide soft semantic supervision over this space through zero-shot distillation.

\section{Method}
\label{sec:method}

The proposed framework learns a video encoder without relying on manually annotated source videos or a predefined source-category space. Its central idea is to derive the supervision space directly from the content of unlabeled videos and then distill instance-specific soft target distributions over this automatically constructed space. To this end, the framework employs two complementary vision-language components with distinct roles. First, an instruction-tuned multimodal model generates open-vocabulary descriptions of the unlabeled videos, from which actions, action-related objects, and action-object interactions are extracted to construct an interpretable pseudo-label vocabulary. Second, a frozen video-language model evaluates the relevance of each pseudo-label to each video and produces a soft target distribution over the resulting vocabulary. A student video encoder is pretrained to reproduce these distributions, thereby learning video representations without human-provided source labels.

For downstream adaptation, the pseudo-label projection head is replaced with a task-specific classification head. The frozen video-language model is then used to generate zero-shot soft targets over the actual target label set, allowing supervised fine-tuning with ground-truth labels to be complemented by VLM-derived supervision. Thus, the proposed framework operates across two distinct label spaces: an automatically discovered pseudo-label space during annotation-free pretraining and a predefined downstream class space during target adaptation. The following subsections describe the construction of the pseudo-label space, zero-shot distillation over this space, and target-label-set-aware fine-tuning with zero-shot distillation, respectively.

\subsection{Pseudo-Label Space Construction from VLM-Generated Captions}
\label{method1}

\begin{figure}[t]
    \centering
    \includegraphics[width=\columnwidth]{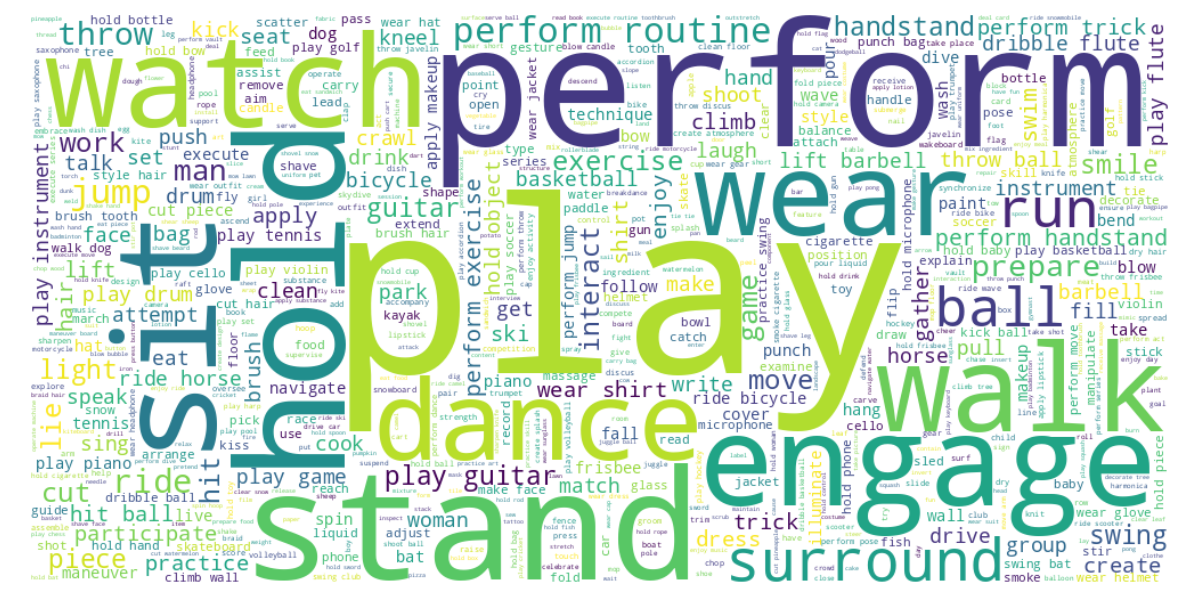}
        \caption{Word-cloud visualization of the constructed pseudo-label space.}
    \label{wordcloud}
\end{figure}

Textual descriptions are first generated for the unlabeled training videos using InternVL2-8B~\cite{internvlref}, an instruction-tuned multimodal large language model (MLLM). For each video, a fixed number of frames is uniformly sampled and provided to the model with the prompt: ``Describe the scene, the objects, and the activity taking place in the video briefly in one or two complete sentences.'' The resulting captions provide open-vocabulary descriptions containing action-related and contextual cues.

The free-form captions are then transformed into a structured pseudo-label vocabulary using the natural language processing framework spaCy~\cite{spacy} with the transformer-based English pipeline \texttt{en\_core\_web\_trf}. The pipeline provides part-of-speech tags, lemmatized forms, and dependency relations for each caption, enabling the extraction of action-related textual units. Specifically, verbs are extracted as action candidates, while nouns are selected as object candidates only when they occur as direct objects of the detected verbs. The corresponding verb--object pairs are also retained to represent action--object interactions. For example, from a caption such as ``a person is playing guitar,'' the procedure extracts \textit{play}, \textit{guitar}, and \textit{play guitar} as candidate pseudo-labels. Thus, the extracted object terms are explicitly conditioned on their dependency relations with actions.

The extracted candidates are aggregated across all captions and filtered in two stages. First, document-frequency constraints are applied to remove terms that occur either too rarely or too frequently across the caption corpus. Second, generic or semantically uninformative words and phrases are removed from the extracted candidate vocabulary using a stop list. Candidate stop-list terms were identified with the assistance of ChatGPT~\cite{chatgpt}, after which the resulting list was manually reviewed and revised. All candidates remaining after these filtering steps are used to define the automatically constructed textual pseudo-label space:

\begin{equation} \mathcal{T}_{\mathrm{}}=\{t_1,t_2,\ldots,t_M\}\label{eq:pseudo_label_space} \end{equation}

where $M$ denotes the number of terms retained after filtering, and each $t_k$ represents a textual pseudo-label corresponding to an action, an action-related object, or an action-object interaction discovered from the generated captions. Fig.~\ref{wordcloud} visualizes the constructed pseudo-label space as a word cloud, with larger terms indicating higher TF-IDF scores.

\subsection{Zero-Shot Distillation over the Pseudo-Label Space}

\begin{figure*}[t]
    \centering
    \includegraphics[width=0.95\textwidth]{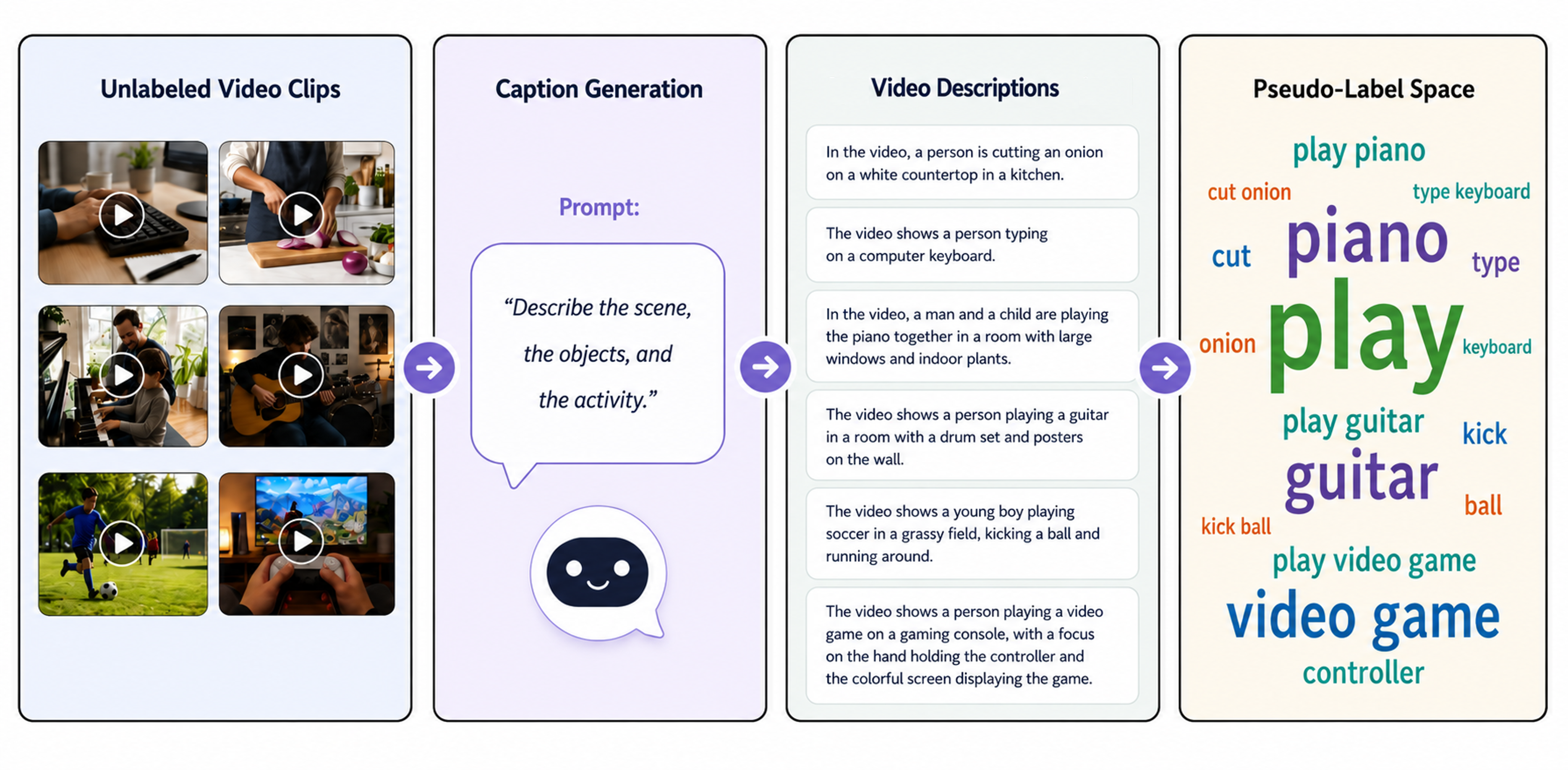}
    \caption{Overview of the pseudo-label space construction pipeline. A vision-language model first generates open-vocabulary descriptions of unlabeled video clips. The captions are then processed and filtered to construct a pseudo-label vocabulary comprising actions, objects, and action--object interactions for annotation-free video pretraining.}
    \label{fig1}
\end{figure*}

\begin{figure*}[t]
    \centering
    \includegraphics[width=0.95\textwidth]{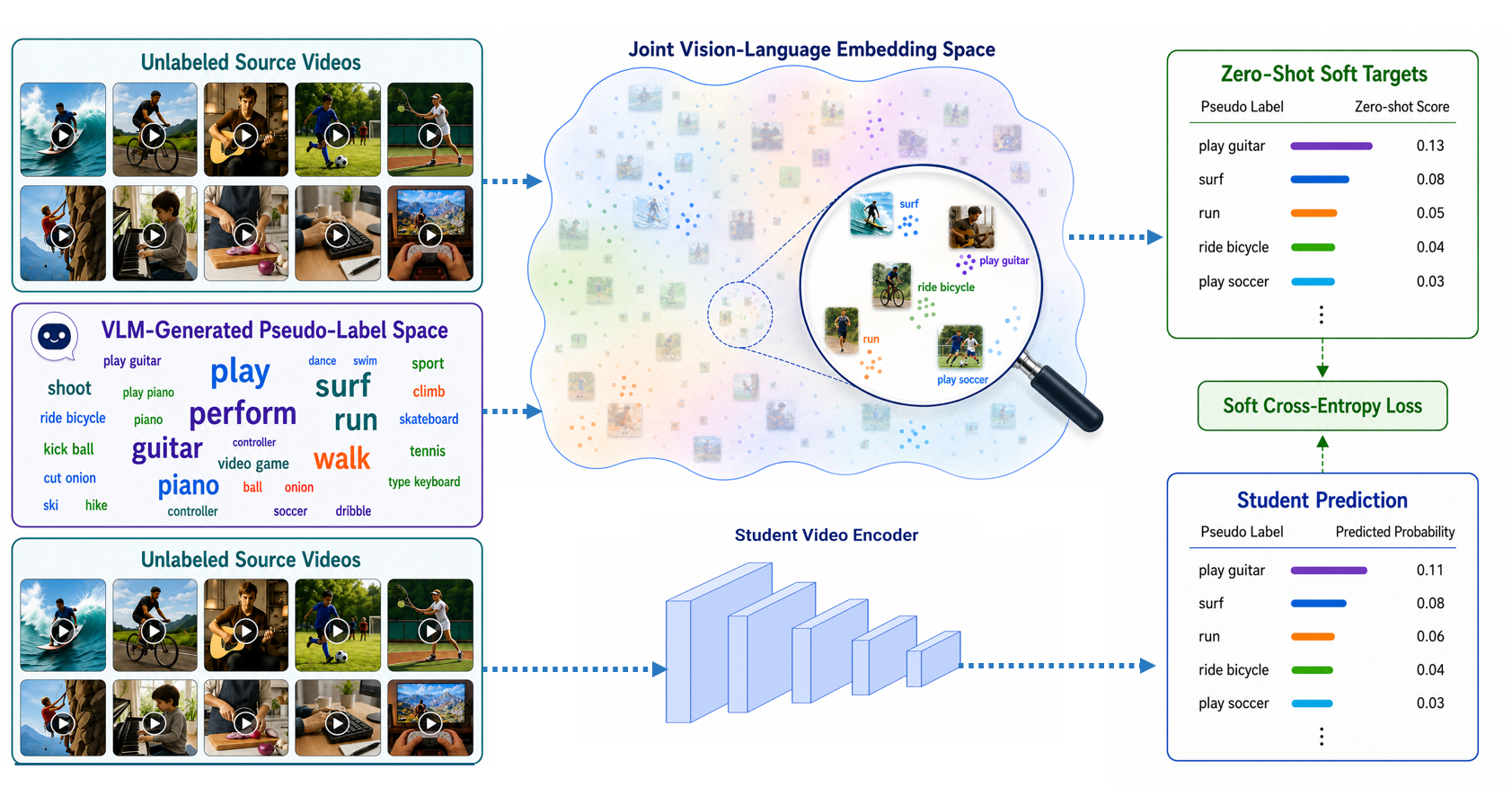}
    \caption{Overview of zero-shot distillation over the caption-derived pseudo-label space. A frozen video-language model compares each unlabeled video with the pseudo-labels in a joint video--text embedding space to generate soft semantic target distributions. A student video encoder is then pretrained to reproduce these distributions, enabling annotation-free representation learning.}
    \label{fig2}
\end{figure*}

After constructing the pseudo-label space $\mathcal{T}$, the frozen InternVideo2-CLIP-S\cite{internvideo} model is used to generate a soft target distribution for each unlabeled video. Rather than assigning a single hard pseudo-label, the teacher measures the compatibility of each video with all textual concepts in $\mathcal{T}$ and produces an instance-specific distribution reflecting their relative relevance. For an unlabeled video $v_i$, let $\phi_V(\cdot)$ and $\phi_T(\cdot)$ denote the frozen video and text encoders of the teacher model, respectively. The video is encoded as $\phi_V(v_i)$, while each pseudo-label $t_k \in \mathcal{T}$ is encoded as $\phi_T(t_k)$. Their compatibility is measured using cosine similarity:

\begin{equation}
s_{i,k} =
\frac{
\phi_V(v_i)^{\top}\phi_T(t_k)
}{
\|\phi_V(v_i)\|
\|\phi_T(t_k)\|}
\label{eq:pseudo_similarity}
\end{equation}

The similarity scores are then converted into a soft distribution over the pseudo-label space by applying a temperature-scaled softmax with temperature parameter $\tau$, which controls the sharpness of the distribution:
\begin{equation}
q_{i,k}^{\mathrm{VLM}} =
\frac{\exp(s_{i,k}/\tau)}
{\sum_{j=1}^{M}\exp(s_{i,j}/\tau)}
\label{eq:pseudo_soft_target}
\end{equation}

A student video encoder $G_{\theta}$ is then trained to match the VLM-derived soft distribution. Given an unlabeled video $v_i$, the student first produces a video representation, which is mapped to the pseudo-label space by a projection head $h_{\psi}(\cdot)$. The resulting student-predicted distribution over the $M$ pseudo-labels, denoted by $p_i^{\mathrm{S}} \in \mathbb{R}^{M}$, is defined as:
\begin{equation}
p_i^{\mathrm{S}} =
\mathrm{softmax}\left(h_{\psi}(G_{\theta}(v_i))\right)
\label{eq:student_prediction_pseudo}
\end{equation}

The annotation-free pretraining objective is formulated as a soft cross-entropy loss between the zero-shot teacher distribution and the student prediction. Let $N_u$ denote the number of unlabeled videos used for pretraining. By minimizing the distillation objective in Eq.~\eqref{eq:zero_shot_distillation_loss}, the student encoder learns to reproduce the soft target distribution derived from video--text similarities computed by the frozen video-language model.

\begin{equation}
\mathcal{L}_{\mathrm{ZSD}} =
-\frac{1}{N_u}
\sum_{i=1}^{N_u}
\sum_{k=1}^{M}
q_{i,k}^{VLM}
\log p_{i,k}^{S}
\label{eq:zero_shot_distillation_loss}
\end{equation}

\subsection{Target-Label-Set-Aware Fine-Tuning with Zero-Shot Distillation}
After annotation-free pretraining, the projection head defined over the VLM-generated pseudo-label space is removed and replaced with a task-specific classification head. The pretrained student encoder is then adapted to the downstream action recognition task using both the labeled and unlabeled target training splits. The labeled target videos provide supervision through hard ground-truth labels, while the unlabeled target videos are used to exploit the zero-shot capability of the frozen video-language teacher over the downstream target label space. In this way, supervised fine-tuning is complemented by VLM-derived soft semantic supervision.

For a target dataset with $C$ downstream action classes, the textual target label set is defined as

\begin{equation}
\mathcal{C}=\{c_1,c_2,\ldots,c_C\}
\end{equation}

where each $c_j$ denotes a natural-language prompt constructed from the corresponding class name, such as ``a video of a person [action].'' Given a target video $v_i$, the frozen video-language model computes a zero-shot similarity score between the video and each class prompt. Using the frozen video and text encoders $\phi_V(\cdot)$ and $\phi_T(\cdot)$, respectively, the similarity between $v_i$ and $c_j$ is computed as

\begin{equation}
s_{i,j} =
\frac{
\phi_V(v_i)^{\top}\phi_T(c_j)
}{
\|\phi_V(v_i)\|
\|\phi_T(c_j)\|}
\label{eq:target_similarity}
\end{equation}

The resulting similarity scores are converted into a soft target distribution over the $C$ downstream action classes using a temperature-scaled softmax with temperature parameter $\tau$:

\begin{equation}
q_{i,j}^{\mathrm{VLM}} =
\frac{\exp(s_{i,j}/\tau)}
{\sum_{m=1}^{C}\exp(s_{i,m}/\tau)}
\label{eqvlmtarget}
\end{equation}

The student prediction over the downstream action classes is obtained using the pretrained video encoder $G_{\theta}$ and a task-specific classification head $g_{\psi}(\cdot)$. The resulting student prediction over the $C$ target classes, denoted by $p_i^{\mathrm{S}} \in \mathbb{R}^{C}$, is defined as

\begin{equation}
p_i^{\mathrm{S}} =
\mathrm{softmax}\left(g_{\psi}(G_{\theta}(v_i))\right).
\label{eq}
\end{equation}

The fine-tuning objective combines supervision from the ground-truth target labels with the zero-shot soft targets produced by the frozen video-language teacher. Let $N_l$ denote the number of labeled target videos, and let $y_{i,j}$ denote the one-hot ground-truth label for class $j$. The supervised cross-entropy loss is defined as

\begin{equation}
\mathcal{L}_{\mathrm{CE}} =
-\frac{1}{N_l}
\sum_{i=1}^{N_l}
\sum_{j=1}^{C}
y_{i,j}\log {p_{i,j}^{\mathrm{S}}} 
\end{equation}

To complement the hard target labels with the relative class relationships encoded by the frozen teacher, the target-aware zero-shot distillation loss is computed over the unlabeled target videos. Let $N_u$ denote the number of unlabeled target videos used during fine-tuning. The loss is defined as

\begin{equation}
\mathcal{L}_{\mathrm{TLSA-ZSD}} =
-\frac{1}{N_u}
\sum_{i=1}^{N_u}
\sum_{j=1}^{C}
q_{i,j}^{\mathrm{VLM}} \log p_{i,j}^{\mathrm{S}}
\label{eq:target_kd_loss}
\end{equation}

Unlike the annotation-free pretraining stage, where the teacher distribution is defined over the automatically generated pseudo-label space $\mathcal{T}$, the distribution in Eq.~\eqref{eqvlmtarget} is defined directly over the downstream action classes in $\mathcal{C}$.

The final fine-tuning objective is given by

\begin{equation}
\mathcal{L}_{\mathrm{FT}} =
\lambda_{\mathrm{ft}}
\mathcal{L}_{\mathrm{CE}}
+
\lambda_{\mathrm{distill}}
\mathcal{L}_{\mathrm{TLSA-ZSD}}
\end{equation}

where $\lambda_{\mathrm{ft}}$ and $\lambda_{\mathrm{distill}}$ control the contributions of the supervised and distillation loss terms, respectively. By minimizing $\mathcal{L}_{\mathrm{FT}}$, the student model learns from the ground-truth target labels while also preserving the embedding structure provided by the frozen video-language teacher. 

\section{Experiments}
\label{sec:experiments}

\begin{table*}[t]
\centering
\caption{Comparison with semi-supervised video action recognition methods on UCF101 and HMDB51 under different labeled-data regimes. Results are reported as video-level Top-1 classification accuracy.}
\label{tab:combined_label}

\begin{threeparttable}

\begin{tabular}{lccccccc}
\toprule

\multirow{2}{*}{\textbf{Method}}
& \multicolumn{2}{c}{\textbf{UCF101}}
& \multicolumn{2}{c}{\textbf{HMDB51}}
& \multirow{2}{*}{\textbf{Distillation}}
& \multirow{2}{*}{\textbf{Modality}}
& \multirow{2}{*}{\textbf{Backbone}} \\

\cmidrule(lr){2-3}
\cmidrule(lr){4-5}

& \textbf{1\%}
& \textbf{10\%}
& \textbf{40\%}
& \textbf{50\%}
& & & \\

\midrule

Supervised
& 8.2 & 24.0~\cite{videossl}\tnote{a} & 18.0~\cite{videossl}\tnote{a} & 30.7~\cite{videossl}\tnote{a}
& \xmark & V & R3D-18 \\

VideoSSL~\cite{videossl}
& -- & 42.0 & 32.7 & 36.2
& \cmark & V & R3D-18 \\

DANet~\cite{danet}
& -- & 64.6 & -- & --
& \cmark & V & R3D-18 \\

CMPL~\cite{cmpl}
& 23.8 & 67.6 & -- & --
& \xmark & V & R3D-18 \\

LTG~\cite{ltg}
& -- & 62.4 & 46.5 & 48.4
& \xmark & V+TG & R3D-18 \\

MvPL~\cite{mvpl}\tnote{b}
& -- & 55.5 & 30.5 & 33.9
& \xmark & V+TG+F & R3D-18 \\

L2A~\cite{l2a}
& -- & 60.1 & 42.1 & 46.3
& \cmark & V & R3D-18 \\

ActorCutMix~\cite{actorcutmix}
& -- & 40.2 & 32.9 & 38.2
& \xmark & V & R(2+1)D-34 \\

FD-VLM~\cite{ours}
& 24.2 & 62.4 & -- & 34.5
& \cmark & V & R3D-18 \\

TimeBalance~\cite{timebalance}
& 29.1 & 69.8 & 49.8 & 51.4
& \xmark & V & R3D-18 \\

\midrule

\textbf{LEViL}
& \textbf{54.3} & \textbf{73.3} & \textbf{51.8} &  \textbf{55.6}
& \cmark & V & R3D-18 \\

\bottomrule
\end{tabular}

\begin{tablenotes}[para,flushleft]
\footnotesize
\item[a] Results are taken from the corresponding cited works.
\item[b] Reimplementation results reported in~\cite{ltg}.
\end{tablenotes}

\end{threeparttable}
\end{table*}

\begin{table*}[t]
\centering
\caption{Comparison of video pretraining strategies for action recognition on UCF101 and HMDB51. Results are reported as video-level Top-1 classification accuracy.}
\small
\scriptsize
\begin{threeparttable}
\begin{tabular}{cccccc}
\toprule
\textbf{Pretrain} & \textbf{Backbone} & \textbf{Strategy} & \textbf{N $\times$ H/W} & \textbf{UCF101} & \textbf{HMDB51} \\
\midrule

None & R3D-18 & None & 16$\times$112 & 42.4\cite{3dstpuzzle}\tnote{a} & 25.3\cite{distinit}\tnote{a} \\

None & R3D-18 & ImageNet Inflation & 16$\times$112 & 74.3\cite{danet}\tnote{a} & -- \\

UCF/HMDB & R3D-18 & DANet\cite{danet} & 8$\times$112  & 76.8 & -- \\

Kinetics+Sports-1M & R(2+1)D-18 & DistInit\cite{distinit} & 32$\times$112 & 85.7 & 54.9 \\
Kinetics+Sports-1M & R(2+1)D-18 & DistInit\cite{distinit} & 8 $\times$112 & -- & 40.3\\

Kinetics+Sports-1M & R3D-18 & DistInit\cite{distinit} & 8 $\times$112 & -- & 39.9  \\

Kinetics & R3D-18 & Supervised Pretraining & 16$\times$112  & 87.8\cite{res}\tnote{a} & 59.3\cite{res}\tnote{a} \\

Kinetics & R3D-18 & Supervised Pretraining & 8$\times$224  &81.7 & 61.2\\

Kinetics & R(2+1)D & Supervised Pretraining & 16$\times$112 & 96.8\cite{cliporder}\tnote{a} & 74.5\cite{cliporder}\tnote{a} \\

Sports-1M & C3D & Supervised Pretraining & 16$\times$112 & 82.3\cite{3drotnet}\tnote{a} & -- \\

Kinetics & R3D-18 & 3DRotNet\cite{3drotnet} & 16$\times$112 & 62.9 & 33.7 \\

MiT & R3D-18 & 3DRotNet\cite{3drotnet} & 16$\times$112 & 62.8 & 29.6 \\

UCF & C3D & PMAS\cite{wang19} & 16$\times$112 & 58.8 & 32.6 \\

Kinetics & C3D & PMAS\cite{wang19} & 16$\times$112& 61.2 & 33.4\\

Kinetics & C3D & 3D ST-Puzzle\cite{3dstpuzzle} & 16$\times$112 & 60.6 & 28.3 \\

Kinetics & R3D-18 & 3D ST-Puzzle\cite{3dstpuzzle} & 16$\times$112& 65.8 & 33.7 \\

UCF & R(2+1)D-18 & ClipOrder\cite{cliporder} & 16$\times$112 & 72.4 & 30.9 \\

UCF & R3D-18 & ClipOrder\cite{cliporder} & 16$\times$112 & 64.9 & 29.5 \\

Kinetics & R(2+1)D-18 & PacePred\cite{pacepred} & 16$\times$112 & 77.1 & 36.6 \\

Kinetics & S3D-G & SpeedNet\cite{speednet} & 64$\times$224 & 81.1 & 48.8 \\

Kinetics & I3D & SpeedNet\cite{speednet} & 64$\times$224 & 66.7 & 43.7 \\

Kinetics & R(2+1)D-18 & VideoMoCo\cite{videomoco} & 32$\times$112 & 78.7 & 49.2 \\

Kinetics & R3D-18 & VideoMoCo\cite{videomoco} & 32$\times$112 & 74.1 & 43.6 \\

UCF & R3D-18 & TCLR\cite{tclr} & 16$\times$112 & 82.4 & 52.9 \\

Kinetics & R3D-18 & TCLR\cite{tclr} & 16$\times$112 & 84.1 & 53.6 \\

Kinetics & R3D-18& CSTP\cite{stor} & 16$\times$112 & 70.5 & 34.4 \\
\midrule
UCF+HMDB+Kinetics\tnote{b} & R3D-18 & LEViL$^{c}$ & 8$\times$224 & 72.5 & 56.2 \\
\bottomrule
\end{tabular}
\begin{tablenotes}[flushleft]
\footnotesize
\item[] 
\textsuperscript{a} Results are taken from the corresponding cited works.
\textsuperscript{b} Uses 5\% subsets of Kinetics-400 and Kinetics-600.
\textsuperscript{c} ZSD pretraining only.

\end{tablenotes}

\end{threeparttable}

\label{maintable}
\end{table*}

\subsection{Datasets}

The annotation-free pretraining stage uses an unlabeled video pool consisting of training videos from UCF101~\cite{ucf} and HMDB51~\cite{hmdb}, together with 5\% subsets of Kinetics-400 and Kinetics-600~\cite{kinetics}.
\footnote{The 5\% subsets of Kinetics-400 and Kinetics-600 used in this study were downloaded from the publicly available Kaggle repository at \url{https://www.kaggle.com/datasets/rohanmallick/kinetics-train-5per} (accessed Jun. 19, 2026).} Although the framework can be scaled to larger unlabeled video collections, doing so would increase the computational cost. The present setting therefore evaluates whether effective semantic supervision can be distilled from a relatively compact pretraining pool. During pretraining, no action labels from any of these datasets are used; all videos are treated solely as unlabeled inputs for VLM-based caption generation, pseudo-label space construction, and zero-shot distillation.

The proposed method is evaluated on the UCF101 and HMDB51 action recognition benchmarks. For the limited-label experiments, the labeled/unlabeled splits and labeled-data ratios are adopted from~\cite{sefar}. Performance is measured by fine-tuning on the labeled training split of each benchmark and evaluating on the corresponding test split. For full-data fine-tuning, the official split 1 of each target dataset is used.

\subsection{Implementation Details}
All experiments are performed on a single NVIDIA GeForce RTX 5090 GPU. \texttt{InternVL2-8B}~\cite{internvlref} is used only for caption generation, where 8 uniformly sampled frames from each video are resized to $448 \times 448$ and provided to the model together with the captioning prompt. Zero-shot similarity scores are computed using the frozen \texttt{InternVideo2-CLIP-S} model~\cite{internvideo}. The pseudo-label vocabulary is constructed using spaCy~\cite{spacy} with its transformer-based English pipeline, \texttt{en\_core\_web\_trf}, followed by TF-IDF filtering with a minimum document frequency of 5 and a maximum document frequency of 0.25. After filtering, the resulting pseudo-label vocabulary contains 2391 textual pseudo-labels. During pretraining and fine-tuning, a single clip of 8 consecutive frames is randomly sampled from each video and resized to $224 \times 224$. During evaluation, 10 temporal clips are sampled deterministically, and video-level predictions are obtained by averaging the clip-level scores.

The student model is implemented as a 3D ResNet-18 in PyTorch~\cite{pytorch} and initialized from scratch with \texttt{weights=None}. During annotation-free pretraining, the student is optimized with AdamW using a learning rate of $10^{-4}$ for 100 epochs, with a batch size of 80. Soft targets are obtained by applying a temperature-scaled softmax to the InternVideo2-CLIP-S similarity scores with $\tau=1/50$. During semi-supervised fine-tuning, each mini-batch contains 40 labeled and 40 unlabeled videos. The supervised and target-aware distillation losses are equally weighted by setting $\lambda_{\mathrm{ft}}=0.5$ and $\lambda_{\mathrm{distill}}=0.5$. During downstream finetuning, the same optimizer and learning rate as in the pretraining stage are used, and all finetuning experiments are conducted for 20 epochs.

\subsection{Experimental Results}

The proposed method is evaluated for downstream action recognition under both semi-supervised and fully supervised settings. Table~1 presents the semi-supervised results on UCF101 and HMDB51 under different labeled-data regimes in terms of video-level Top-1 classification accuracy. In this setting, the complete proposed framework is applied, including annotation-free pretraining followed by target-label-set-aware zero-shot distillation during downstream fine-tuning. Therefore, the reported results reflect the overall performance of the two-stage framework. To provide additional context, Table~1 also indicates whether each method uses distillation, together with its input modality and backbone architecture. The supervised baselines correspond to training the video model from scratch using only the available labeled subset and without any additional supervision, and therefore serve as lower-bound references for assessing the benefit of the compared semi-supervised methods under the same labeled-data regime.

The compared methods represent several directions in semi-supervised video action recognition. VideoSSL~\cite{videossl}, one of the earliest semi-supervised action recognition frameworks, transfers knowledge from an image-based teacher to a video student, while DANet~\cite{danet} extends this paradigm by incorporating multiple teachers and contrastive objectives. LTG~\cite{ltg} and MvPL~\cite{mvpl} incorporate motion-related cues through temporal gradients and optical flow, respectively. L2A~\cite{l2a} and ActorCutMix~\cite{actorcutmix} focus on augmentation strategies for semi-supervised training. TimeBalance~\cite{timebalance} employs spatial and temporal teachers trained with self-supervised pretext tasks and dynamically balances their predictions according to the input video. CMPL~\cite{cmpl}, in contrast, uses a primary backbone together with a lightweight auxiliary network with a different architectural design. These architecturally distinct networks learn complementary representations and generate pseudo-labels for each other through cross-model pseudo-labeling. Overall, the compared methods span several major directions, including teacher--student learning, motion-based supervision, augmentation-based training, and self-supervised pretext tasks. To provide a more controlled comparison, Table~1 is restricted to methods using CNN-based video backbones; the transformer-based methods SVFormer~\cite{svformer} and SeFAR~\cite{sefar} are therefore excluded.

FD-VLM~\cite{ours} is the most closely related method because it also employs video--text multimodal supervision. However, FD-VLM relies on feature-level distillation followed by fine-tuning with hard ground-truth labels. In such a setting, the transferred knowledge remains embedded in a high-dimensional feature space, making it difficult to explicitly inspect or control the semantic information distilled to the student. Moreover, because VLM-derived supervision is not explicitly retained during downstream fine-tuning, the model may gradually shift toward the limited hard-label supervision, increasing the risk of overfitting in low-label regimes. In contrast, the proposed method performs pretraining over an automatically constructed pseudo-label space composed of interpretable textual concepts. During downstream adaptation, target-label-set-aware zero-shot distillation further complements the hard labels with soft targets defined over the actual action classes, allowing VLM-derived semantic guidance to be maintained throughout fine-tuning.

As shown in Table~1, LEViL achieves the best performance among all compared methods across every evaluated label regime on both UCF101 and HMDB51. These results demonstrate the effectiveness of the proposed framework under varying levels of label availability, with particularly strong gains in the extremely low-label setting.

Table~2 presents the fully supervised fine-tuning results, where all labeled training videos of the target dataset are used. In this setting, the proposed annotation-free pretraining strategy is evaluated solely as a weight initialization mechanism, without target-label-set-aware zero-shot distillation. The learned initialization is compared with alternative initialization strategies for video models. The results on UCF101 and HMDB51 demonstrate that the representations learned during annotation-free pretraining provide an effective initialization.

Overall, LEViL consistently improves upon supervised training from scratch under limited-label settings and provides transferable representations under full-data fine-tuning. Importantly, these results are obtained without using the complete Kinetics datasets for pretraining. Only 5\% subsets of both Kinetics-400 and Kinetics-600 are incorporated into the unlabeled pretraining pool. The results therefore indicate that transferable video representations can be learned from a comparatively modest unlabeled video collection, supporting the proposed method as a practical alternative to full-scale supervised video pretraining.

\section{Discussion}
\label{sec:discussion}

Label-efficient training of video models has become increasingly important because video annotation is costly and difficult to scale. Compared with images, videos contain temporal information that increases both annotation effort and training complexity. As a result, video models trained from scratch with limited data are more susceptible to overfitting, motivating the use of alternative supervision sources that can exploit unlabeled videos.

The proposed framework addresses this problem by combining annotation-free pretraining with target-label-set-aware fine-tuning. During pretraining, VLM-generated captions are converted into an interpretable pseudo-label space, and the student learns from soft distributions over this space. During downstream adaptation, zero-shot classification is performed over the actual target classes, and the resulting soft targets complement the available hard labels. In this formulation, the distilled knowledge is represented through a textual vocabulary rather than being embedded in a high-dimensional feature space, making the transferred information easier to inspect and control. In addition, the frozen VLM is used only during training for caption generation and soft-target construction. At test time, predictions are produced solely by the student video model, introducing no additional inference cost.

The results are also notable because the proposed method learns transferable video representations from a relatively small unlabeled pretraining pool. Despite using substantially less source data than full-scale video pretraining, the complete framework improves performance under limited-label settings, while the annotation-free pretraining stage alone provides an effective initialization for full-data fine-tuning.
 Since the framework does not depend on source annotations or a predefined source label space, it can also be extended to larger or domain-specific unlabeled video collections.

The proposed framework also has aspects that can be further improved. Since the pseudo-label space is constructed from VLM-generated captions, its quality depends on the quality of these captions. If the generated descriptions fail to capture relevant video content, the extracted pseudo-labels may be less informative. In this work, the pseudo-label construction process is kept simple and general; however, this may not be sufficient for all datasets or application domains. More refined prompt design and pseudo-label selection strategies may therefore improve the quality of the generated supervision. Since the framework is modular, stronger captioning and video-language models can also be incorporated as they become available, potentially improving its performance.
\section{Conclusion}
\label{sec:conclusion}

This work presented an annotation-free video pretraining framework that leverages the joint video--language embedding space of VLMs to construct pseudo-label supervision, together with a target-label-set-aware fine-tuning strategy. The proposed method constructs a textual pseudo-label space from captions generated for unlabeled videos and pretrains a student video encoder by distilling zero-shot soft distributions produced by a frozen video-language model. During downstream adaptation, supervised learning is complemented by target-label-set-aware zero-shot distillation over the actual action label set.

Experiments on UCF101 and HMDB51 demonstrate the effectiveness of the proposed framework under both limited-label and full-label settings. The results show that transferable video representations can be learned without manually annotated source labels or predefined source action categories, even from a relatively small unlabeled pretraining pool. Overall, the proposed method provides a scalable alternative to conventional supervised video pretraining by learning transferable representations without manually annotated source videos.

\section*{Acknowledgment}
The author acknowledges the use of ChatGPT and Gemini for language editing and readability improvements, as well as for assistance in generating and visually refining the schematic illustrations in Figs.~\ref{fig1} and~\ref{fig2}. ChatGPT was also used to assist with the identification of candidate stop-list terms during the pseudo-label space construction described in Section~\ref{method1}. All generative AI-assisted content was reviewed and verified by the author.
\bibliographystyle{IEEEtran}
\bibliography{references}

\end{document}